\documentclass[letterpaper]{article}

\usepackage{natbib,alifeconf}  
\usepackage{url,hyperref,cleveref}
\usepackage{booktabs}

\usepackage{lipsum}

\newcommand\blfootnote[1]{%
  \begingroup
  \renewcommand\thefootnote{}\footnote{#1}%
  \addtocounter{footnote}{-1}%
  \endgroup
}

\title{Limbomorphs}

\author{
    Alex Alvarez$^{1,*}$ and
    Michael Levin$^{2,3}$ \\
    \mbox{}\\
    $^1$ Department of Brain and Cognitive Sciences, Massachusetts Institute of Technology, Cambridge, MA, USA \\
    $^2$ Allen Discovery Center, Tufts University, Medford, MA, USA \\
    $^3$ Wyss Institute for Biologically Inspired Engineering, Harvard University, Boston, MA, USA \\
    $^*$Corresponding author: \texttt{alexrez@mit.edu}
}

\begin{document}

\maketitle

\begin{abstract}
    Artificial life systems are typically defined by a set of dynamical rules over an environment, an agent, or both, from which lifelike patterns may emerge. Gifbreeder is an animated version of the interactive evolutionary computation (IEC) platform Picbreeder, and was initially created to generate visual art. Instead of encoding the agent or the environment, Gifbreeder genomes encode a spatiotemporal field and evolve through the user’s aesthetic selection. The evolved expressions can sometimes resemble motile lifelike creatures that we term \textit{Limbomorphs}, given that they exist in a deterministic three-second looping “limbo”. We assess their behavior via input-space perturbations and find species-specific reactions to different kinds of perturbations. We discuss whether these reactions may reflect goal-directed behavior like navigation, or merely the appearance of it, and more broadly how agent-like dynamics may emerge in a system with no explicitly defined agent, environment, or interaction rules.

\end{abstract}


Submission type: \textbf{Late Breaking Abstract}\\

Data/Code available at: \url{https://github.com/calvarez0/limbomorphs}
\blfootnote{\textcopyright  2026 Alex Alvarez and Michael Levin. Published under a Creative Commons Attribution 4.0 International (CC BY 4.0) license.}

\section{Gifbreeder}
The space of all possible three-second videos is combinatorially enormous. Most artifacts in this space are visually meaningless, and a random sample almost always returns complete noise. Picbreeder is a collaborative IEC platform in which users guide the evolution of images by repeatedly selecting aesthetically preferred variants from a population generated through CPPN-NEAT \citep{secretan2011picbreeder, stanley2007compositional, stanley2004competitive}. This allows for search over vast image space to be far more efficient, often finding recognizable patterns and objects in only a few generations. Similar ideas have been explored in IEC more broadly \citep{dawkins1987blind, takagi2001interactive}, and more recently, it has been shown that time-indexed IEC of compositional pattern-producing networks (CPPNs) can extend static images to support the breeding of 2D and 3D animations \citep{tweraser2018querying}.

Gifbreeder is a web-based IEC platform inspired by Picbreeder that uses NEAT-based search \citep{stanley2002evolving} over CPPN representations to evolve animated GIFs through user-guided selection. The chosen CPPN inputs for each pixel are its coordinates, radial distance from center, time, and a bias constant. The three outputs are interpreted and displayed as HSV channels. Formally, a genome parameterizes a CPPN
\[
F_{\theta}(x, y, d, t, b) \to (h, s, v),
\]
where $\theta$ denotes the network topology, activation functions, and connection weights, $x$ and $y$ the pixel coordinates, $d = \sqrt{x^2 + y^2}$, $t$ the time step, and $b$ the bias node. $x$ and $y$ inputs are normalized between $-0.5$ and $0.5$ while $t$ is normalized between $0$ and $1$.

Phenotypes are generated by querying $F_\theta$ over a pixel grid and t = 45 timesteps, corresponding to 3 seconds at 15 frames per second. Besides the time input, Gifbreeder differs from Picbreeder in its set of activation functions, which were roughly chosen for their apparent ubiquity in nature \citep{turing1952chemical, murray1981pre, french1992development, kondo2010reaction}. Gifbreeder defines a spacetime field rather than a simulated world. This distinction is central to the interpretation of the results below.

\section{Limbomorphs}

\begin{figure}[ht]
    \centering
    \includegraphics[width=3.05in]{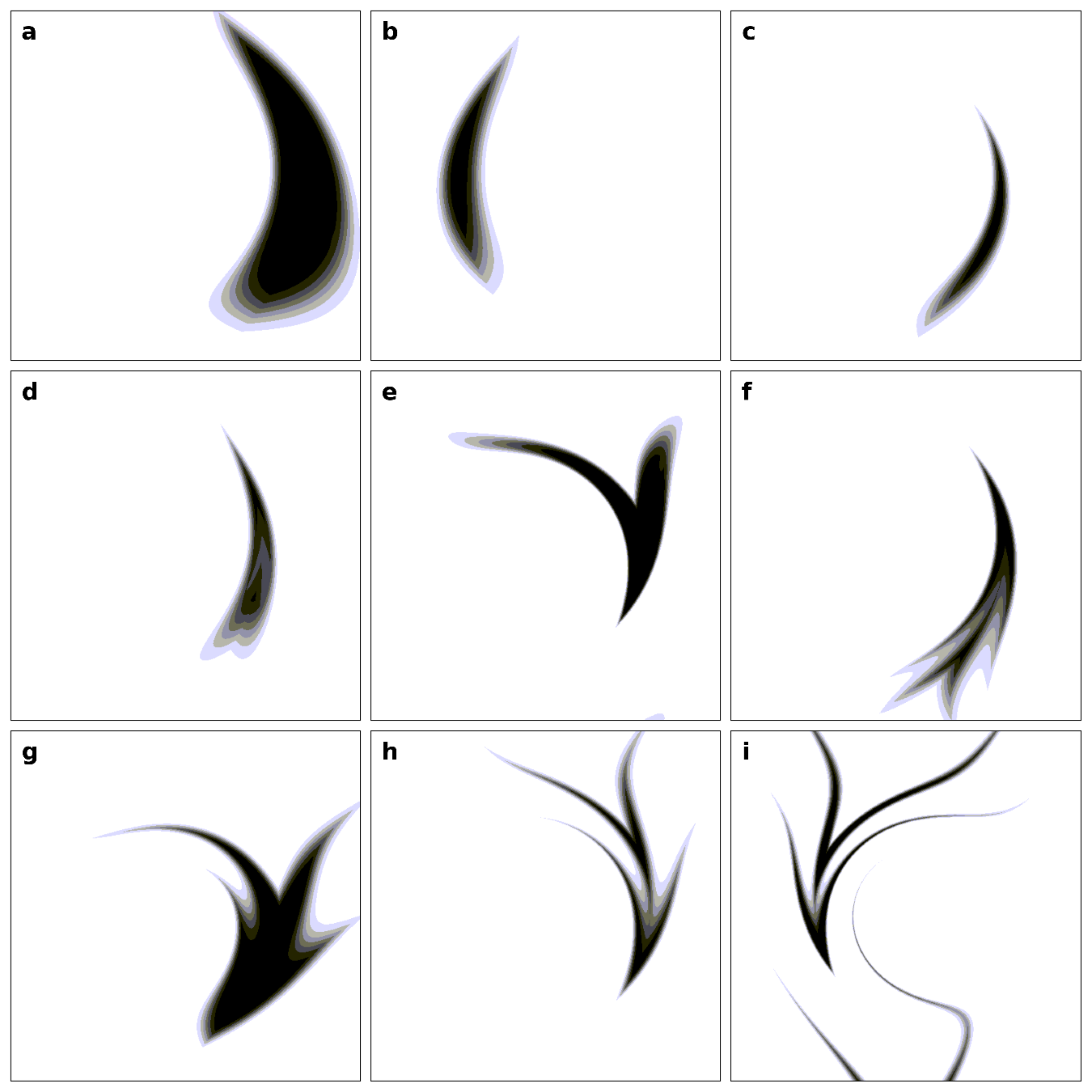}
    \caption{
        Nine species of the same family with increasing morphological complexity. More species can be found at
https://alexrez.com/openendedness/gifbreeder/drawing.html
    }
    \label{fig1}
\end{figure}

The emergence of lifelike beings in Gifbreeder was completely unintended and unexpected since it was initially created to be an art tool. At first, Gifbreeder patterns were desolate and abstract. Occasionally and increasingly, expressions began to appear recognizable and organic. After several evolutionary runs, we began to notice expressions that exhibited apparent animacy \citep{heider1944experimental}. Once a limbomorph was found, it could be used as a branch-off point to evolve a family of creatures. Using this technique, we quickly generated an entire families of limbomorphs with varying degrees of morphological and behavioral complexity (Figure~\ref{fig1}). 

\begin{figure}[b!]
    \centering
    \includegraphics[width=3.25in]{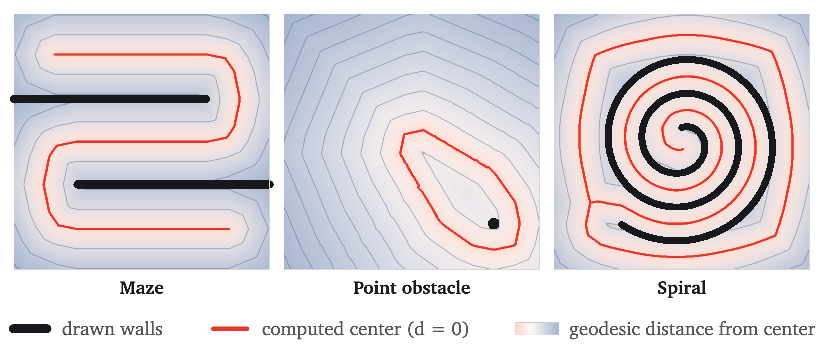}
    \caption{
        Drawn strokes (black) become walls; we compute each region's center (red, d = 0) and feed the geodesic distance from it (shading, contours) to the CPPN as the $d$ input.
    }
    \label{fig2}
\end{figure}

\section{Reactions to perturbations}
Gifbreeder genomes encode the entire GIF and not simply the limbormoph, making it impossible to excise the being from its environment. This makes assessing whether there is any "being" at all particularly challenging. In Lenia \citep{chan2019lenia}, for example, creatures have kernels which act as a receptive field that can inform the creature about incoming obstacles \citep{cool2026agnosiophobia}. Limbomorphs don't have local rules or receptors in analogous ways, however, one way to think about interacting with limbomorphs is to warp the inputs of its CPPN's in a meaningful way. This was the motivation for the \textit{drawing tool}, which warps how $d$ is calculated (Figure~\ref{fig2}). When the user draws walls with the perturbation tool, $d$ is replaced by the geodesic distance from the computed center \textit{ridge}.

\begin{figure}[t!]
    \centering
    \includegraphics[width=3.25in]{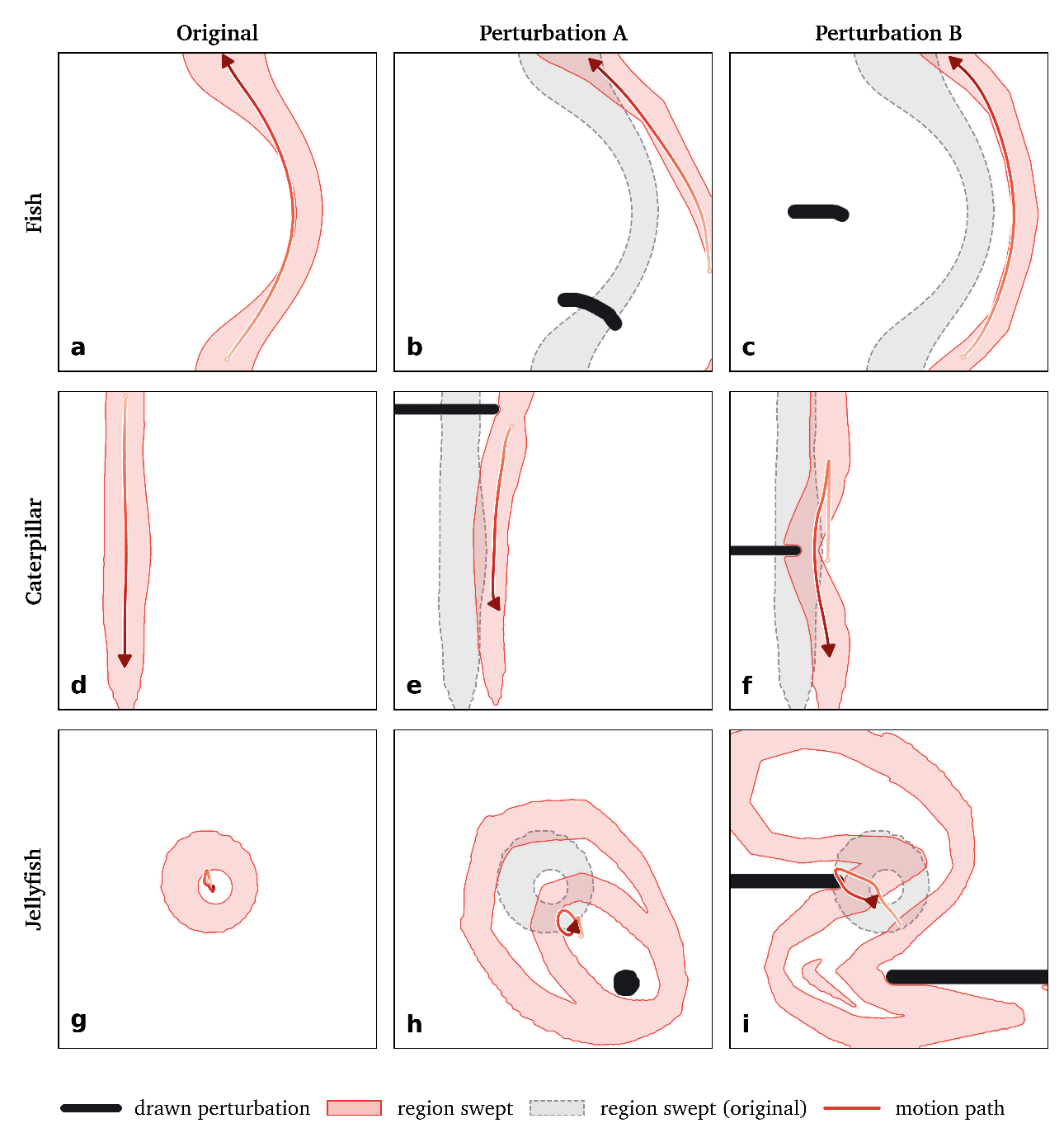}
    \caption{
        Each limbomorph's motion is affected differently depending on the species and the point of perturbation.
    }
    \label{fig3}
\end{figure}

Perturbations were applied to existing expressions using this drawing tool resulting in species-specific behavioral reactions. One limbomorph species, “Fish”, is extremely avoidant to the perturbations (Figure ~\ref{fig3}a-c). This mirrors the “agnosiophobia” behavioral response recently discovered in Lenia, characterized by avoidance of regions of space where no sensory information is available. Another species, "Caterpillar", navigates \textit{away} from the perturbation in the upper regions (Figure ~\ref{fig3}e), but \textit{towards} the perturbation in lower regions (Figure ~\ref{fig3}f). Lastly, creating a "point obstacle" in the bottom right causes the "Jellyfish" species, which normally maintains its topology near the center of the space, to stretch and envelop the perturbation (Figure ~\ref{fig3}h). Astonishingly, when a small maze is drawn, Jellyfish orients its body around the maze (Figure ~\ref{fig3}i).

\section{Conclusion and Discussion}
We present Gifbreeder and its drawing tool as a new medium for studying artificial life. Species-specific reactions to input-space perturbations—like Jellyfish maintaining its topology while repositioning around the maze—suggest that behaviors suggestive of spatial competence and goal-directedness \citep{Levin2019, Heylighen2023} can arise in a system with no explicit representation of agent, environment, or interaction rules. Could limbomorphs provide a useful model system for investigating how basal cognition can emerge from field dynamics alone? Further analysis is needed to determine the mechanisms underlying these behaviors.

\section{Acknowledgements}
We would like to thank Ken Stanley and Yoonsuck Choe for the helpful feedback and encouragement to pursue this idea.

\footnotesize
\bibliographystyle{apalike}
\bibliography{example} 

\end{document}